\documentclass{article}

\usepackage[final]{neurips_2025}

\usepackage[utf8]{inputenc} 
\usepackage[T1]{fontenc}    
\usepackage{hyperref}       
\usepackage{url}            
\usepackage{booktabs}       
\usepackage{amsfonts}       
\usepackage{nicefrac}       
\usepackage{microtype}      
\usepackage{xcolor}         
\usepackage{amsmath}
\usepackage{algorithm}
\usepackage{algpseudocode}
\usepackage{graphicx}
\usepackage{caption}

\title{Experience-Efficient Model-Free Deep Reinforcement Learning Using Pre-Training}

\author{
  Ruoxing (David) Yang \\
  Department of Computer Science \\
  Georgetown University \\
  \texttt{ry216@georgetown.edu} \\
}

\begin{document}

\maketitle

\begin{abstract}
We introduce PPOPT - Proximal Policy Optimization using Pretraining, a novel, model-free deep-reinforcement-learning algorithm that leverages pretraining to achieve high training efficiency and stability on very small training samples in physics-based environments. Reinforcement learning agents typically rely on large samples of environment interactions to learn a policy. However, frequent interactions with a (computer-simulated) environment may incur high computational costs, especially when the environment is complex. Our main innovation is a new policy neural network architecture that consists of a pretrained neural network middle section sandwiched between two fully-connected networks. Pretraining part of the network on a different environment with similar physics will help the agent learn the target environment with high efficiency because it will leverage a general understanding of the transferrable physics characteristics from the pretraining environment. We demonstrate that PPOPT outperforms baseline classic PPO on small training samples both in terms of rewards gained and general training stability. While PPOPT underperforms against classic model-based methods such as DYNA DDPG, the model-free nature of PPOPT allows it to train in significantly less time than its model-based counterparts. Finally, we present our implementation of PPOPT as open-source software, available at github.com/Davidrxyang/PPOPT.
\end{abstract}

\section{Introduction}

Reinforcement learning models traditionally rely on a large amount of environment interactions (experience) to learn a policy for completing some task. However, frequent interactions with a complex computer-simulated environment may incur high computational costs, especially when the environment involves simulating real-world physics. We are also motivated to minimize environment interactions when the environment exists in the real world, such as with robotics applications, as opposed to within a virtual computer simulation. This is because real-world consequences of environment interactions such as mechanical fatigue are often more costly than those incurred by computer simulations. For this study, we will only consider computer-simulated environments where the only scarce resources are computational cost and training time.

When a reinforcement learning agent lacks sufficient experience, its policy may fail to converge reliably. This problem is worsened with the popularization of deep reinforcement learning models such as DQN ~\cite{mnih2013playingatarideepreinforcement} and PPO~\cite{schulman2017proximalpolicyoptimizationalgorithms}, which use deep neural networks to represent the policy. Modeling a policy function using a deep (multi-layer) neural network usually requires more training data than simple tabular methods such as tabular Q-Learner due to the large number of parameters within a neural network, all of which demand a large number of optimization iterations through a large amount of data to converge on an appropriate value. 

We choose to address the classic problem of experience-efficient deep reinforcement learning by answering one simple question: \textit{how do we maximize performance under tight computational cost bounds and sparse experience?} We propose PPOPT - Proximal Policy Optimization using Pretraining - a novel, model-free deep reinforcement learning model designed based on classic proximal policy optimization (PPO). We modify PPO by introducing a new policy network architecture. This new network consists of a pretrained neural network middle section sandwiched between two fully-connected networks. We use classic PPO to pretrain a basic fully-connected neural network policy on an environment with similar physics to the target environment. We then export this pretrained network and insert it into the center of the PPOPT policy network. This helps the agent learn the target environment with high efficiency because it will leverage a general understanding of the transferrable physics characteristics from the pretraining environment. 

Through experimenting with different pretraining and target environment combinations, hyperparameters, and policy network design, we find that PPOPT consistently outperforms baseline PPO under limited experience. We find that PPO is generally unable to converge onto any sensible policy under very small experience samples. PPOPT, on the other hand, consistently discovers practical semi-sensible policies very quickly by taking advantage of its general understanding of environment physics gained from the pretraining process. The pretraining section also adds a powerful stabilizing effect to the PPOPT training cycle. Compared to the erratic learning curve produced by PPO under small experience samples, the PPOPT learning curve exhibits much greater stability and shows a clear upwards trend. 

While PPOPT, in its current iteration, already consistently outperforms model-free deep reinforcement learning methods such as PPO which do not leverage pretraining, it is still unable to outperform model-based deep reinforcement learning methods such as DYNA DDPG ~\cite{moerland2022modelbasedreinforcementlearningsurvey}. While model-based methods achieve higher performance under the same amount of experience, these methods require significantly more training time and computational resources. PPOPT provides a practical balance between performance and training time under sparse experience scenarios. 

\section{Related Work}

\subsection{Pretraining for Deep Reinforcement Learning}

Other recent efforts have also explored pretraining as a strategy to reduce the heavy sample requirements in deep reinforcement learning. Among the most comprehensive overviews of this trend is the survey by Xie et al.~\cite{xie2022pretrainingdeepreinforcementlearning}, which categorizes pretraining techniques along several dimensions including the nature of data used (e.g., expert demonstrations, offline trajectories, unlabeled observations) and the specific components of the reinforcement learning pipeline being pretrained (e.g., representation encoder, policy, value function, world model).

\textbf{Representation Pretraining.} This approach aims to learn generalizable features from raw observations using unsupervised or self-supervised learning, which are later reused during policy learning. Such methods include contrastive learning, predictive modeling, or masked reconstruction objectives. By decoupling feature extraction from policy optimization, agents require fewer interactions to start improving their behavior in the target task. This approach is most similar to the approach we use for PPOPT.

\textbf{Policy and Value Pretraining.} Here, the policy and/or value networks are initialized using supervised learning on expert demonstrations or from related tasks. This accelerates early learning and reduces exploration requirements. Techniques such as behavior cloning or inverse reinforcement learning often serve as the foundation.

\textbf{World Model Pretraining.} Some methods construct forward dynamics or generative models trained on offline datasets. These models are then used to simulate trajectories for planning or policy updates. Although not novel on their own, when pretrained separately, these components can dramatically improve training efficiency and enable zero-shot or few-shot generalization across environments. This approach combines pretraining and model-based reinforcement learning, which will be discussed in the next related-works subsection.

Xie et al. emphasize that pretraining is particularly advantageous in scenarios with sparse rewards, complex visual observations, or costly simulations. Their analysis also points to challenges, such as representation overfitting, distribution mismatch between pretraining and downstream environments, and the lack of standardized benchmarks for evaluating pretraining effectiveness in reinforcement learning.

Our work builds on these insights by designing a pretraining approach tailored specifically to policy networks operating in physics-based environments, leveraging their shared structure to improve learning stability and experience efficiency in sparse experience scenarios.

\subsection{Model-Based Reinforcement Learning}

Model-based reinforcement learning has emerged as a promising approach to enhance experience efficiency by leveraging learned models of environment dynamics for planning and policy optimization. Moerland et al.~\cite{moerland2022modelbasedreinforcementlearningsurvey} provide a comprehensive survey of model-based reinforcement-learning methods, categorizing them based on model learning strategies and planning-learning integration techniques.

\textbf{Dynamics Model Learning.} The survey discusses various approaches to learning environment dynamics, including deterministic and probabilistic models, as well as methods to handle stochasticity, uncertainty, and partial observability. Accurate dynamics models enable agents to simulate future trajectories, reducing the need for extensive real-world interactions.

\textbf{Planning and Learning Integration.} Moerland et al. categorize planning-learning integration into three main strategies: (1) planning with a known model, (2) learning both the model and the policy, and (3) planning over a learned model without explicit policy learning. They analyze how these strategies impact sample efficiency and performance.

\textbf{Implicit Model-Based RL.} The survey also introduces the concept of implicit model-based reinforcement-learning, where model learning and planning are integrated into end-to-end learning frameworks. These approaches aim to combine the benefits of model-based and model-free methods, potentially improving both efficiency and robustness.

\section{Preliminaries}
\subsection{Proximal Policy Optimization (PPO)}

Proximal Policy Optimization (PPO)~\cite{schulman2017proximalpolicyoptimizationalgorithms} is a widely adopted policy gradient method that balances the stability of trust-region policy optimization (TRPO) with the simplicity of first-order optimization. It serves as the baseline algorithm in our work and also the architectural foundation from which we derive our proposed approach, PPOPT.

PPO optimizes a clipped surrogate objective that penalizes large deviations between the new policy and the old policy, thereby enforcing a soft trust region. Let $\pi_\theta$ denote the current policy parameterized by $\theta$, and $\pi_{\theta_{\text{old}}}$ be the behavior policy used to collect data. For a given trajectory sampled from $\pi_{\theta_{\text{old}}}$, PPO maximizes the following clipped objective:

\[
\mathcal{L}^{\text{CLIP}}(\theta) = \mathbb{E}_t \left[ \min\left( r_t(\theta) \hat{A}_t, \text{clip}(r_t(\theta), 1 - \epsilon, 1 + \epsilon) \hat{A}_t \right) \right],
\]
where $r_t(\theta) = \frac{\pi_\theta(a_t|s_t)}{\pi_{\theta_{\text{old}}}(a_t|s_t)}$ is the probability ratio, and $\hat{A}_t$ is the estimated advantage function. The clipping function prevents the new policy from diverging too far from the old one, stabilizing training and reducing policy oscillation.

PPO typically alternates between sampling trajectories from the environment and performing multiple epochs of minibatch updates using the surrogate objective. This approach improves sample efficiency and reduces variance in gradient estimates.

\vspace{1em}
\noindent\textbf{Pseudocode: Proximal Policy Optimization (PPO)}

\begin{algorithm}[H]
\caption{Proximal Policy Optimization (PPO)}
\begin{algorithmic}[1]
\State Initialize policy parameters $\theta$ and value function parameters $\phi$
\For{each iteration}
    \State Collect trajectories $\mathcal{D} = \{(s_t, a_t, r_t, s_{t+1})\}$ using $\pi_\theta$
    \State Compute advantages $\hat{A}_t$ using GAE
    \State Compute returns-to-go $\hat{R}_t$
    \For{epoch = $1$ to $K$}
        \For{each minibatch $\mathcal{B} \subset \mathcal{D}$}
            \State Compute probability ratio: $r_t(\theta) = \frac{\pi_\theta(a_t|s_t)}{\pi_{\theta_{\text{old}}}(a_t|s_t)}$
            \State Compute clipped objective: \\
            \hskip1em $\mathcal{L}_t^{\text{CLIP}} = \min\left(r_t(\theta)\hat{A}_t,\ \text{clip}(r_t(\theta), 1 - \epsilon, 1 + \epsilon)\hat{A}_t\right)$
            \State Compute value loss: $\mathcal{L}_t^{\text{VF}} = (V_\phi(s_t) - \hat{R}_t)^2$
            \State Perform gradient descent on: \\
            \hskip1em $\mathcal{L} = \mathbb{E}_t\left[ \mathcal{L}_t^{\text{CLIP}} - c_1 \mathcal{L}_t^{\text{VF}} + c_2 \mathcal{H}[\pi_\theta](s_t) \right]$
        \EndFor
    \EndFor
\EndFor
\end{algorithmic}
\end{algorithm}

\vspace{1em}
\noindent PPO’s modular architecture and stable performance make it an ideal candidate for modification. In this work, we propose PPOPT, which extends PPO by introducing a pretrained middle section in the policy network architecture. This pretrained module enables the reuse of transferable physics representations, enhancing experience efficiency and generalization in sparse experience scenarios. 

\subsection{DYNA DDPG}

DYNA DDPG combines the Deep Deterministic Policy Gradient (DDPG) algorithm~\cite{lillicrap2019continuouscontroldeepreinforcement} with the DYNA-style model-based reinforcement learning framework~\cite{sutton1990integrated}. In this hybrid approach, an explicit dynamics model is learned alongside the actor and critic networks. The model is then used to generate synthetic rollouts that augment the replay buffer, effectively increasing the amount of experience available for learning.

This integration improves sample efficiency by enabling policy updates from both real and imagined trajectories. In practice, the agent alternates between collecting environment data, updating the dynamics model, and performing off-policy learning using both real and synthetic transitions. While DYNA DDPG can achieve strong performance, it requires careful balancing of model accuracy and rollout depth to avoid compounding error and instability.

We include DYNA DDPG as a baseline in our experiments to benchmark against model-based methods. Its access to synthetic experience highlights the trade-off between experience efficiency and computational cost compared to our model-free approach, PPOPT.

\section{Gymnasium: Multi-Joint Dynamics with Contact} 

To maintain consistency across experiments and produce a generalizable implementation, we select environments from the Multi-Joint Dynamics with Contact (MuJoCo) physics engine simulator environment library under the Gymnasium library~\cite{towers2024gymnasiumstandardinterfacereinforcement}, developed by OpenAI. MuJoCo is designed for fast and accurate simulation of physical systems involving complex bodies that make physical contact with the environment. It is particularly popular in reinforcement learning research because it supports high-speed simulation, accurate modeling of contacts (both soft and hard), and differentiable continuous action and observation spaces, which are useful for gradient-based optimization and control. 

\subsection{Pretraining: Inverted Pendulum}

We choose the Inverted Pendulum environment as the pretraining environment. This environment upgrades the classic cartpole environment by incorporating the MuJoCo physics engine. The task is to balance a pole on top of a cart. The pole is attached to the cart on the bottom end and free-moving on the top end. The cart can be moved linearly left or right in order to keep the pole upright. We choose this environment as the pretraining environment due to its simplicity. Compared to complex MuJoCo environments such as Ant or Humanoid which involve complex bodies with multiple movable joints, the Inverted Pendulum environment only has one controllable dimension. As the simplest environment that utilizes the MuJoCo physics engine, Inverted Pendulum is a good candidate for pretraining because it is computationally cheap. 

\subsection{Inverted Double Pendulum}

We choose Inverted Double Pendulum as the first sparse experience target environment. Inverted Double Pendulum is closely related to Inverted Pendulum as it simply modifies Inverted Pendulum by adding a second free-moving pole to the top of the first pole. However, this simple change greatly increases the control difficulty of the body as well as complicating the observation space. We choose this environment due to its close relation to the pretraining environment.

\subsection{Hopper}

We choose Hopper as the second target environment. Hopper relation to the pretraining environment, Inverted Pendulum, is much more abstract. They share a vague pole-like characteristic, as well as the MuJoCo physics engine. The hopper is a two-dimensional one-leg figure with four body parts: a torso, a thigh, a bottom leg, and a foot. The goal is to make hops in the forwards direction. The joints are controlled by applying torque. We choose this environment to examine the effect of transferring pretrained skills to an environment with similar physics characteristics but significantly different action control scheme.

\section{Technical Approach: Architecture}

PPOPT replaces PPO's basic policy neural network with a novel neural network architecture. The novel architecture consists of three components: the input network, the pretrained network, and the output network.

\begin{figure}[h]
    \centering
    \includegraphics[width=0.8\linewidth]{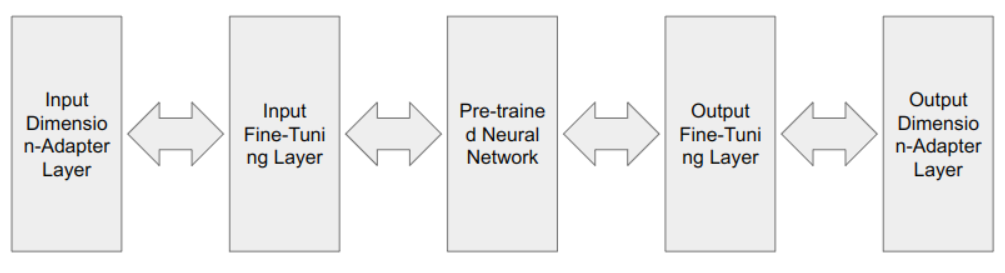}
    \caption{PPOPT network architecture}
    \label{fig:PPOPT_network}
\end{figure}

\subsection{Pretrained Network}

The pretrained network is a fully connected neural network with four linear layers. The input and output layers are mapped to the state and action dimensions of the pretraining environment. These layers sandwich two hidden layers with 128 neurons each. This design follows the model of a traditional deep reinforcement learning policy neural network. We train a fresh copy of this network with randomly initialized parameters using classic PPO in the pretraining environment. We then transplant this network into the PPOPT architecture by exporting the parameters from the pretraining agent policy network and loading them into the pretrained section of the PPOPT network.

\subsection{Input Network}

The input network is attached to the front of the pretrained network, connecting the state space of the environment to the input side of the pretrained network. The first layer in the input layer is the input dimension-adapter layer. This layer is mapped to the dimensionality of the target environment state space. The main goal for this layer is to learn a transformation between the target environment state space dimensionality and the pretrained environment state space dimensionality which is utilized by the pretrained network input side. 

The second layer in the input network is the input fine-tuning layer. This layer follows the dimensionality of the pretraining environment state space and so matches the dimensionality of the pretrained network input layer. We choose to match dimensionality to facilitate the pretrained network transplant process. 

\subsection{Output Network}

The output network follows a similar design philosophy as that of the input network. The output side of the pretrained network is attached to the output fine-tuning layer. The output fine-tuning layer matches the dimensionality of the pretraining environment's action space. 

The output fine-tuning layer feeds into the output dimension-adapter layer. This layer matches the dimensionality of the target environment action space and serves, in part, to learn a transformation from the action dimension of the pretrained environment to that of the target environment. 

\section{Technical Approach: Training}

The full PPOPT agent is trained in two phases: pretraining and main training. 

\subsection{Pretraining}

Before directly training the PPOPT model, we first separately train the pretrained section of the PPOPT network. We use the standard PPO algorithm to train a policy network in the selected pretraining environment. This PPO policy network is identical to the pretrained network section of the PPOPT model. After pretraining, we export the parameters of the PPO policy network and load these parameters into the pretrained section of the PPOPT network. 

We choose to discard the trained value network produced by the pretraining process. While it makes sense to pretrain a policy network to retain learned physics properties among similar environments, the value network does not have comparable shareable attributes. The reward value definition of each environment and task is very different and hard to transfer, so pretraining the PPOPT value network does not provide sigificant utility. 

\subsection{Main Training}

The main training loop uses the standard PPO algorithm to train on the target environment and runs stochastic gradient ascent throughout all sections of the PPOPT network. Instead of freezing the weights of the pretrained section to fully preserve the pretrained skills, we allow the trained weights of the pretrained section to vary normally. We intend for the learning process to reshape the skills learned during the pretraining process to better adapt those skills generated from the pretraining environment to the novel target environment. We choose a lower learning rate for optimizing the pretrained section because we would still like to preserve the features learned by the pretraining process partially. 

Furthermore, the extra parameters introduced by the pretrained section induce a stabilizing side-effect to the training process. The large amount of parameters compared to classic PPO makes it more difficult for the algorithm to generate drastic shifts in behavior, which helps mitigate the risk of common issues such as catastrophic forgetting.

\section{Algorithm}

\begin{algorithm}
\caption{PPOPT: Proximal Policy Optimization using Pretraining}
\begin{algorithmic}[1]
\Require Pretraining environment $\mathcal{E}_{\text{pre}}$, target environment $\mathcal{E}_{\text{target}}$, pretraining iterations $N_{\text{pre}}$, training iterations $N_{\text{train}}$
\Ensure Trained policy $\pi_{\theta'}$ for $\mathcal{E}_{\text{target}}$

\State Initialize policy network $\pi_\theta$ with parameters $\theta$
\For{$i = 1$ to $N_{\text{pre}}$}
    \State Collect trajectories from $\mathcal{E}_{\text{pre}}$ using $\pi_\theta$
    \State Estimate advantages $\hat{A}_t$
    \State Compute PPO objective $L^{\text{CLIP}}(\theta)$
    \State Update $\theta$ using gradient ascent on $L^{\text{CLIP}}$
\EndFor

\State Extract pretrained parameters $\theta_{\text{core}}$ from $\theta$

\State Construct new policy $\pi_{\theta'}$ with:
\State \hskip1em Input dimension-adapter layer
\State \hskip1em Input fine-tuning layer
\State \hskip1em Pretrained core layers with $\theta_{\text{core}}$
\State \hskip1em Output fine-tuning layer
\State \hskip1em Output dimension-adapter layer
\State Initialize new layers' parameters randomly

\For{$j = 1$ to $N_{\text{train}}$}
    \State Collect trajectories from $\mathcal{E}_{\text{target}}$ using $\pi_{\theta'}$
    \State Estimate advantages $\hat{A}_t$
    \State Compute PPO objective $L^{\text{CLIP}}(\theta')$
    \State Update $\theta'$ using gradient ascent on $L^{\text{CLIP}}$
\EndFor

\State \Return $\pi_{\theta'}$
\end{algorithmic}
\end{algorithm}

\section{Experiments}

We demonstrate the performance of PPOPT by running experiments on the Double Inverted Pendulum and Hopper environments. For both experiments, we select Inverted Pendulum as the pretraining environment. We select 3e-4 as the initial dynamically decreasing pretraining learning rate. We pretrain over 600 episodes. We compare the results of PPOPT with baseline PPO and DYNA DDPG. All models are trained on a machine with an 8th gen intel processor. We include a separate chart excluding DYNA DDPG to highlight the improvement of PPOPT over baseline PPO. 

To account for the randomness inherent to training learning agents, we execute all experiments 5 times and take average statistics to use in our analysis. We also include the range of experiment results as the shaded section. 

\subsection{Experiment 1: Double Inverted Pendulum}

For Double Inverted Pendulum, we select 3e-4 as the initial learning rate for PPOPT's input and output networks, baseline PPO, and DYNA DDPG for comparison consistency. We choose 1e-4 as the pretrained core learning rate for PPOPT. We train all models over 200 episodes. 

Average Training times in Seconds: 
PPO: 30.23
DYNA DDPG: 151.25
PPOPT: 34.54

\begin{figure}[h]
    \centering
    \includegraphics[width=0.7\linewidth]{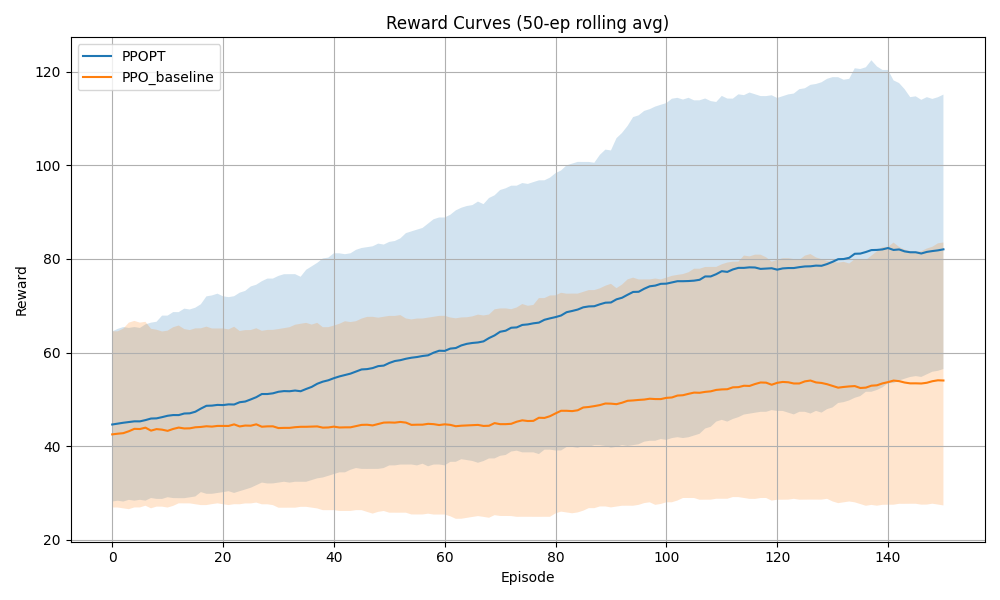}
    \caption{Results from Double Inverted Pendulum, excluding DYNA DDPG}
    \label{fig:EXP1_NODYNA}
\end{figure}

\begin{figure}[h]
    \centering
    \includegraphics[width=0.7\linewidth]{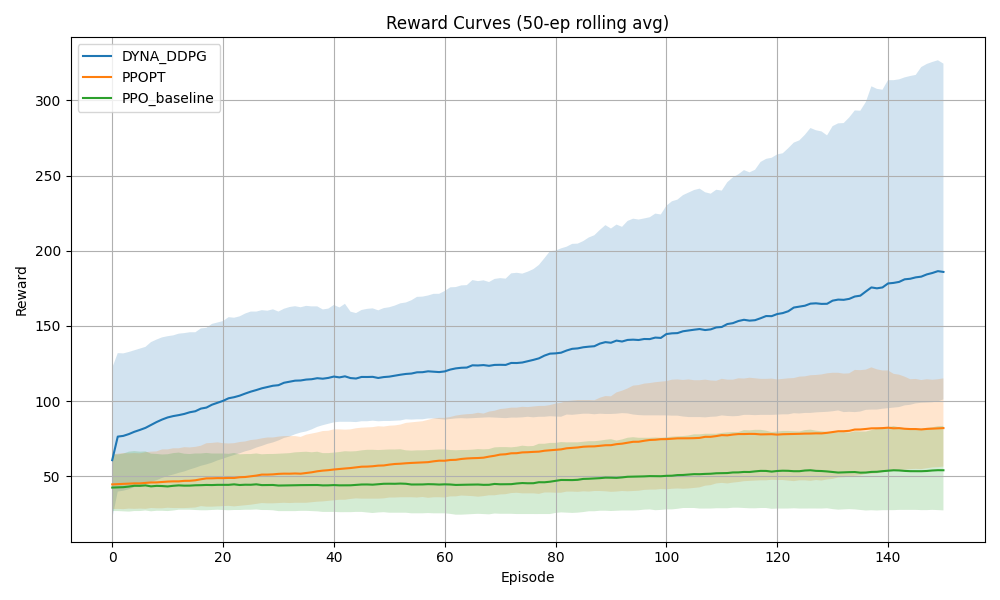}
    \caption{Results from Double Inverted Pendulum}
    \label{fig:EXP1}
\end{figure}

\subsection{Experiment 2: Hopper}

For Hopper, we select 3e-4 as the initial learning rate for PPOPT's input and output networks, baseline PPO, and DYNA DDPG for comparison consistency. We choose 1e-5 as the pretrained core learning rate for PPOPT. We train all models over 200 episodes. 

PPO displayed significant instability on Hopper. To produce a clear graph, we modified the baseline PPO results by setting all highly negative rewards (smaller than -10) to -10. 

Average Training times in Seconds: 
PPO: 26.73
DYNA DDPG: 1307.22
PPOPT: 44.82

\begin{figure}[h]
    \centering
    \includegraphics[width=0.7\linewidth]{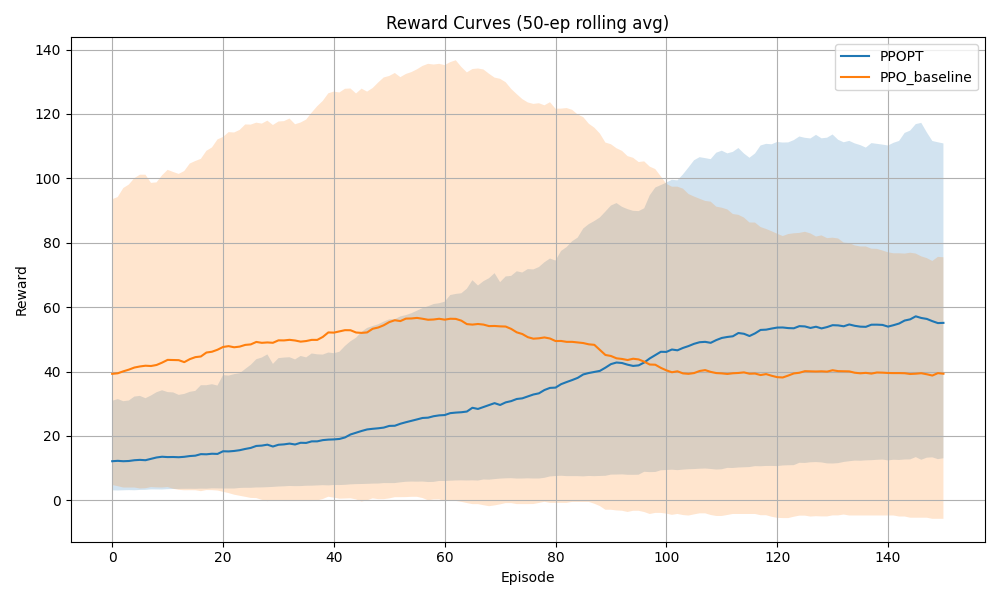}
    \caption{Results from Hopper, excluding DYNA DDPG}
    \label{fig:EXP1_NODYNA}
\end{figure}

\begin{figure}[h]
    \centering
    \includegraphics[width=0.7\linewidth]{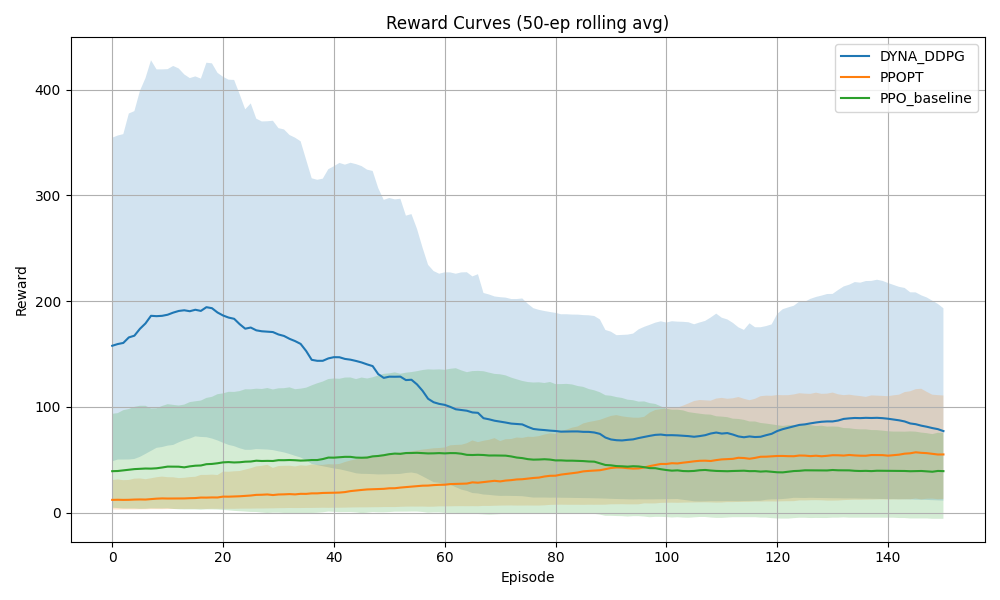}
    \caption{Results from Hopper}
    \label{fig:EXP1}
\end{figure}

\section{Analysis}

From experimental results, we determine that PPO is generally unable to converge on any sort of sensible policy within such sparse experience scenarios. Furthermore, PPO also exhibits great instability during these periods because it needs more training episodes to converge. 

Since PPOPT enjoys the advantage of starting with a base pretrained section, it exhibits much greater stability compared to PPO under sparse experience scenarios. From both experients, PPOPT is also able to achieve higher performance than PPO. 

DYNA DDPG outperforms PPO and PPOPT in both experiment scenarios. However, DYNA DDPG takes signicantly more time to train compared to PPOPT and PPO because generating simulated experience and training on this simulated experience is time and computational resource intensive. This relationship highlights the tradeoff between training time and performance under sparse experience scenarios. 

We recognize that, while it is unusual for PPO to converge with such sparse experience, it is still possible for PPO to find a sensible policy very quickly due to the stochastic nature of parameter initialization and training. However, PPOPT generally exhibits greater stability under such scenarios and finds sensible approximate policies with much greater reliability. 

\section{Conclusion}
We address the problem of experience scarcity in deep reinforcement learning by proposing Proximal Policy Optimization using Pretraining (PPOPT), a novel model-free deep reinforcement learning algorithm that leverages pretraining on similar environments to maximize training efficiency under sparse experience scenarios in a complex target environment. PPOPT introduces a new policy network architecture that involves an input network section, a pretrained network section, and an output network section. We demonstrate that PPOPT consistently outperforms traditional PPO under sparse-experience scenarios. We show that while PPOPT does not achieve the same high performance as DYNA DDPG, it consumes significantly less training time. 

\section{Future Work}
We are interested in exploring combining PPOPT and DYNA to produce a model-based version of PPOPT. We would aim to strike a balance between training time and performance by limiting the amount of training cycles allocated for the simulated experience to reduce the high training time of model-based methods, while leveraging PPOPT to maximize the utility of the limited simulated experience training cycles. 

We are also interested in deploying PPOPT at larger scale, exploring how PPOPT design principles can be used to benefit other areas of AI research such as LLM design.

\bibliographystyle{unsrt}
\bibliography{references}

\end{document}